\documentclass[nohyperref]{article}

\usepackage{microtype}
\usepackage{graphicx}
\usepackage{subcaption}
\usepackage{booktabs} 
\usepackage{algorithm}
\usepackage{multicol}
\usepackage{makecell}

\usepackage{hyperref}

\usepackage[accepted]{icml2023}

\usepackage{amsmath}
\usepackage{amssymb}
\usepackage{mathtools}
\usepackage{amsthm}

\usepackage[capitalize,noabbrev]{cleveref}

\theoremstyle{plain}

\theoremstyle{definition}

\theoremstyle{remark}

\usepackage[textsize=tiny]{todonotes}

\icmltitlerunning{~ \hfill CLAM: Selective Clarification for Ambiguous Questions with Generative Language Models \hfill}

\begin{document}

\twocolumn[
\icmltitle{CLAM: Selective Clarification for Ambiguous Questions\\ with Generative Language Models}



\icmlsetsymbol{equal}{*}

\begin{icmlauthorlist}
\icmlauthor{Lorenz Kuhn}{equal,oatml}
\icmlauthor{Yarin Gal}{oatml}
\icmlauthor{Sebastian Farquhar}{oatml}

\end{icmlauthorlist}

\icmlaffiliation{oatml}{OATML Group, Department of Computer Science, University of Oxford}

\icmlcorrespondingauthor{Lorenz Kuhn}{lorenz.kuhn@cs.ox.ac.uk}

\icmlkeywords{Machine Learning, ICML}

\vskip 0.3in
]



\printAffiliationsAndNotice{}  

\begin{abstract}
Users often ask dialogue systems ambiguous questions that require clarification. We show that current language models rarely ask users to clarify ambiguous questions and instead provide incorrect answers. To address this, we introduce CLAM: a framework for getting language models to selectively ask for clarification about ambiguous user questions. In particular, we show that we can prompt language models to detect whether a given question is ambiguous, generate an appropriate clarifying question to ask the user, and give a final answer after receiving clarification. We also show that we can simulate users by providing language models with privileged information. This lets us automatically evaluate multi-turn clarification dialogues. Finally, CLAM significantly improves language models' accuracy on mixed ambiguous and unambiguous questions relative to SotA. 

\end{abstract}

\section{Introduction}

\begin{figure}[]
    \centering
    \begin{subfigure}[b]{0.45\textwidth}
    \centering
        \includegraphics[width=0.8\textwidth]{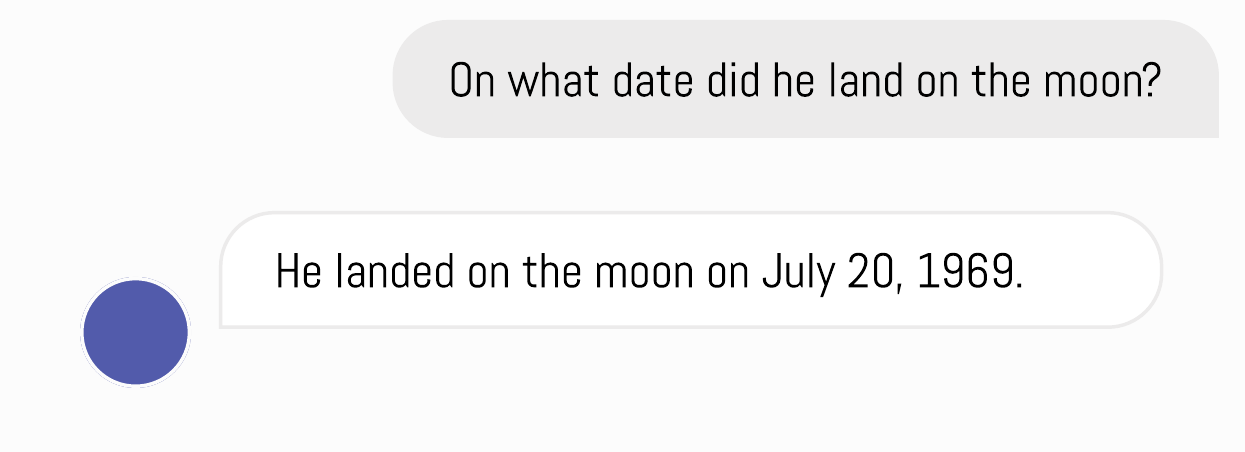}
        \vspace{-4mm}
        \caption{Normal LM behaviour.}
    \end{subfigure}\hfill
    \begin{subfigure}[b]{0.45\textwidth}
    \centering
        \includegraphics[width=0.8\textwidth]{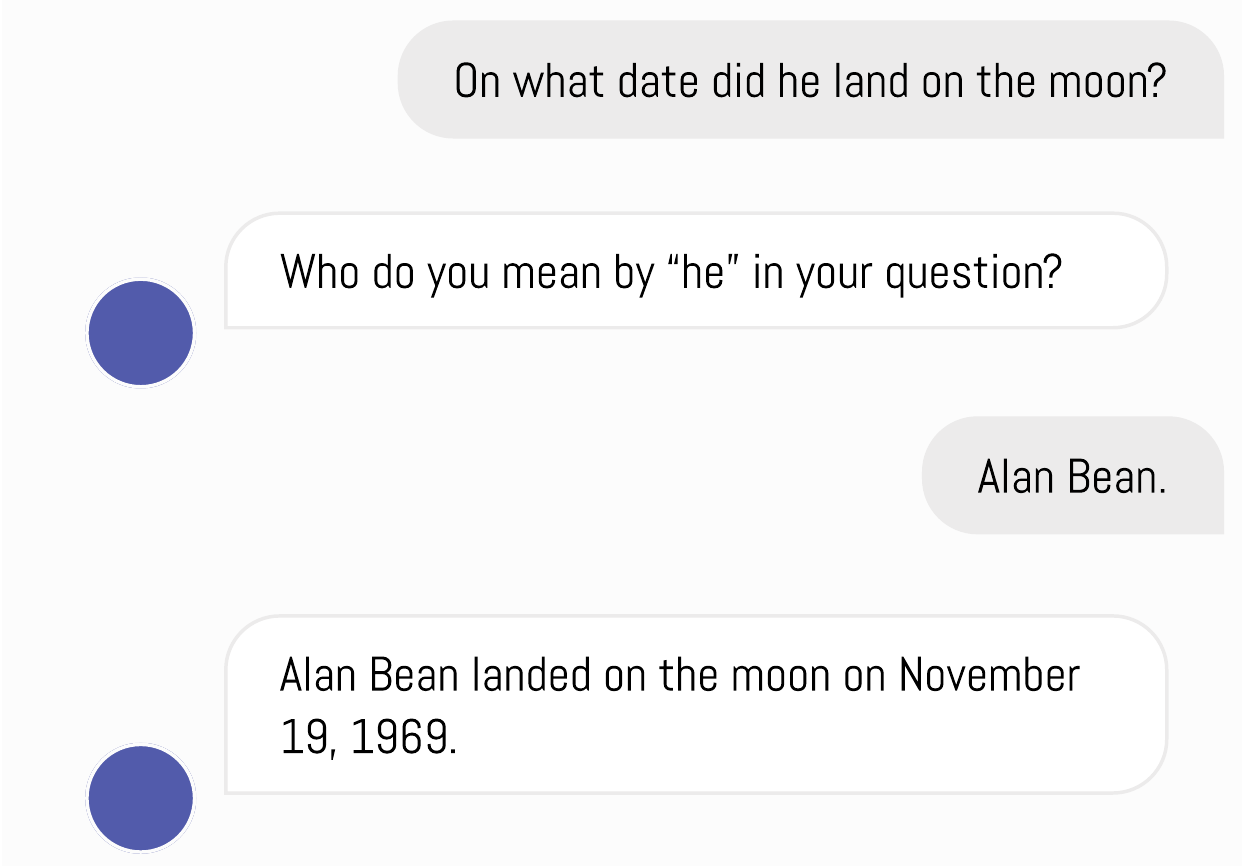}
        \vspace{-4mm}
        \caption{CLAM (ours)}
    \end{subfigure}
    \caption{(a) Normally, LMs answer one of many interpretations given an ambiguous question. (b) Our method uses few-shot classification to detect ambiguous questions and selectively asks for clarifying information needed to answer the question.}%
    \label{figure_before_and_after}
\end{figure}

Recent Transformer-based large language models (LLMs) are often accurate on open- and closed-book question-answering tasks \citep{chung2022scaling, hoffmann2022training}. These data sets typically consist of well-defined questions with enough information to have a unique answer. As these language models are deployed, however, they will often face \textit{vague} user questions.
A user will have some well-defined question in mind but accidentally pass an under-specified question to the question-answering model. For example, a user might want to ask ``On what date did Alan Bean land on the moon?'' but accidentally pass the question ``When did he land on the moon?'' to the language model, as illustrated in Figure \ref{figure_before_and_after}.  The fact that user requests are often ambiguous is well-established in the information retrieval literature (see \citet{keyvan2022approach} for an overview) but has so far received little attention in the LLM question-answering community. This is despite widespread reports of models ``hallucinating'' responses when faced with unanswerable questions (see \citet{ji2022survey} for an overview). In this paper, we show that state-of-the-art language models rarely ask for clarification about ambiguous user inputs and thus perform poorly when answering ambiguous questions.

\begin{figure}
    \centering
    \includegraphics[scale=0.2]{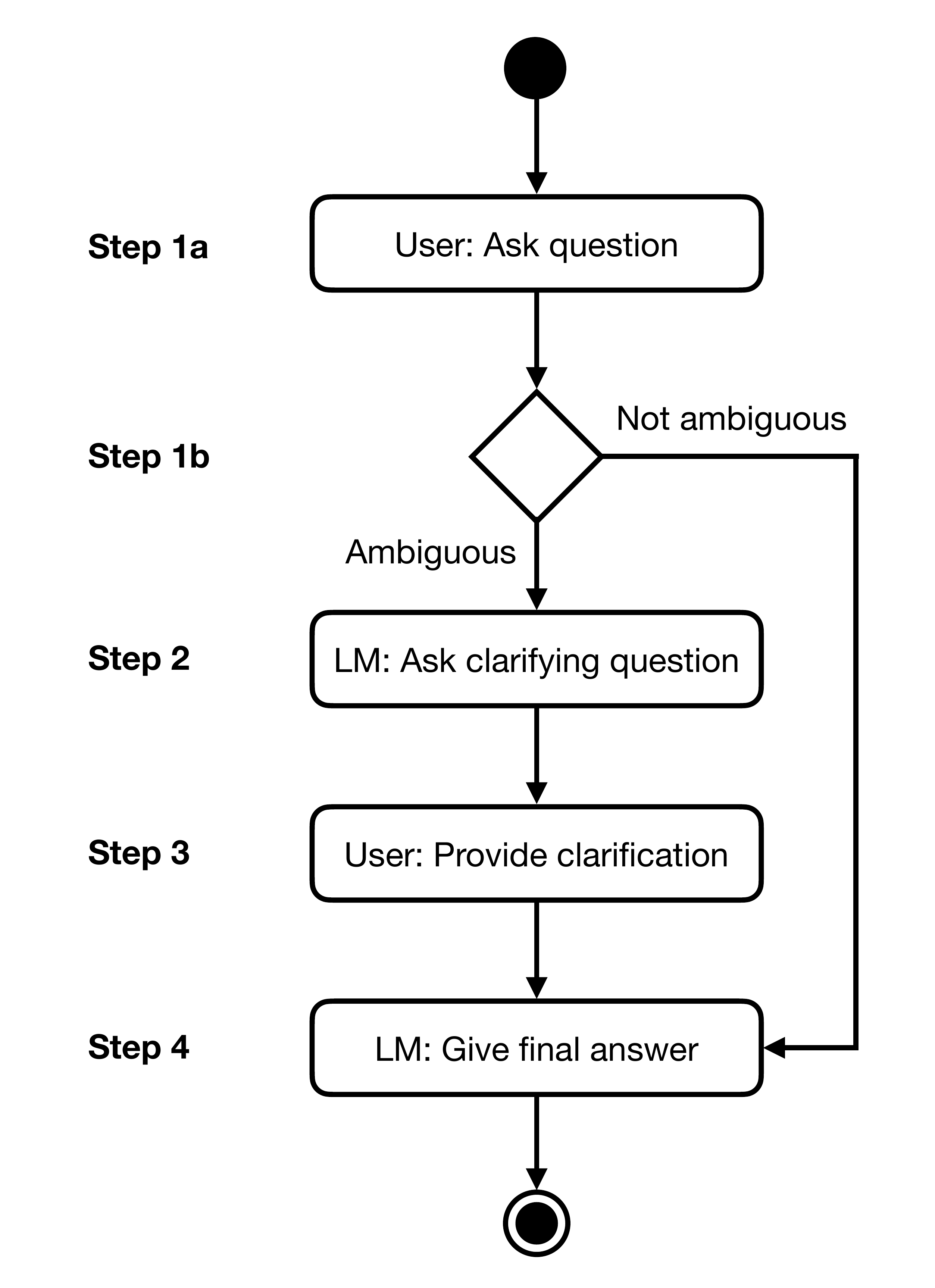}
    \caption{\textbf{Overview of Selective Clarification}}
    \label{figure_selective_clarification}
\end{figure}

\begin{figure*}
    \centering
    \includegraphics[scale=0.2]{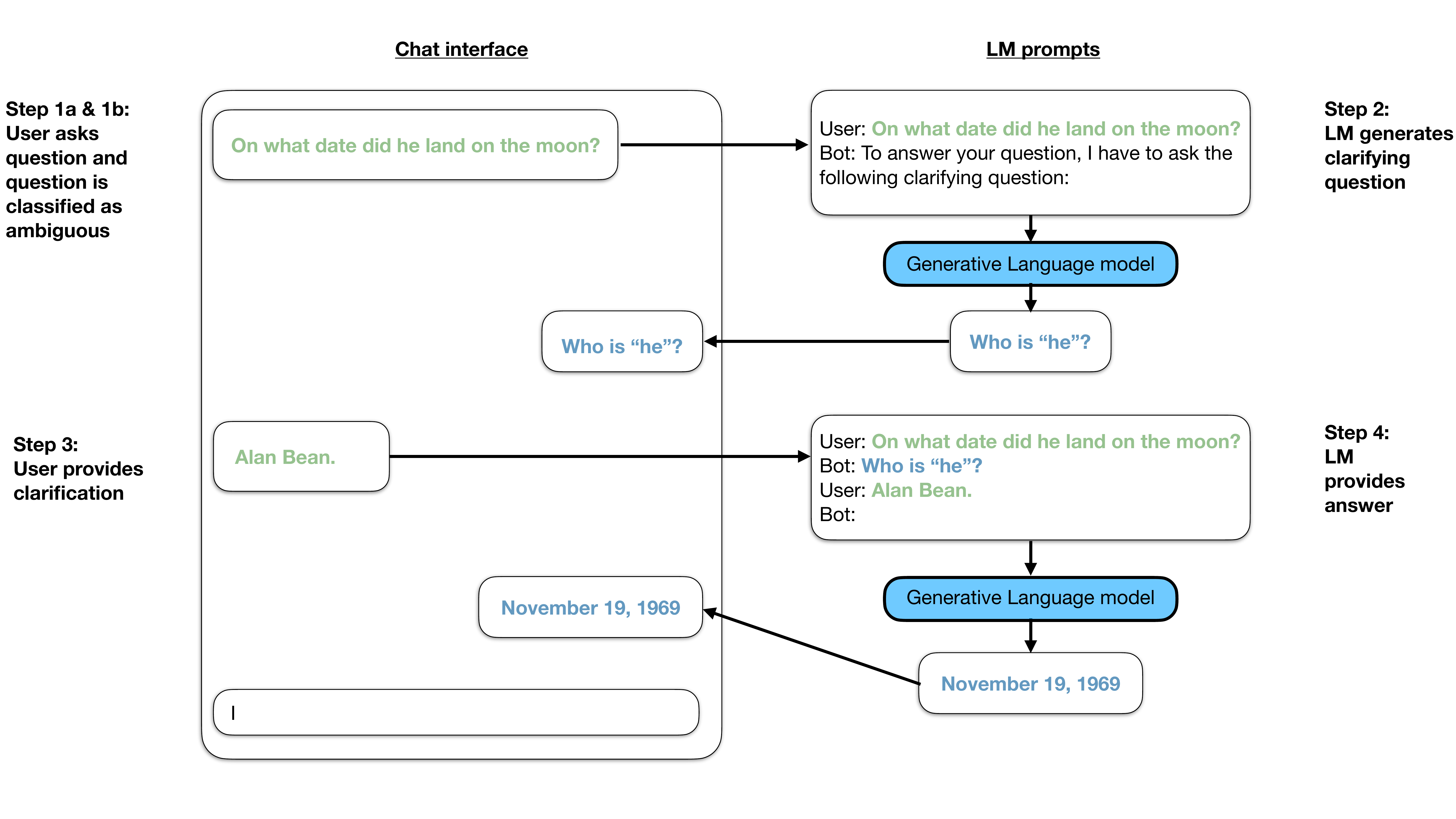}
    \caption{\textbf{Overview of the prompts used to clarify ambiguous user inputs}. In step 1a, the user asks an ambiguous question. In step 1b (omitted for clarity) the question is classified as ambiguous using a few-shot prompt. In step 2, the model is few-shot prompted to generate a clarifying question about the ambiguous user input. In step 3, the user (or an oracle model, see Section \ref{section_automatic_evaluation_protocol}) provides clarifying information given the clarifying question. In step 4, the model is prompted to answer the initial question given the clarification from the user.}
    \label{figure_prompting_diagram}
\end{figure*}

To address this issue, we introduce CLAM, a framework that can significantly improve language models' question-answering performance in a setting we describe as \textbf{selective clarification question answering}.
The framework involves: identifying ambiguous questions, prompting the model to resolve ambiguity, and answering the disambiguated question.
In this paper, we demonstrate some of the tools that can be used to implement this procedure, showing that even a relatively simple implementation can greatly improve performance. We also show that this method reliably only asks for clarification when the user input is actually ambiguous and thus avoids asking unnecessary clarifying questions.

Conceptually, CLAM can be seen as a form of \textbf{meta-cognition}---often described as thinking-about-thinking \citep{lai2011metacognition}. That is, CLAM is a way of improving model performance by prompting the model to explicitly ``reason'' about a property of the given problem before trying to solve it. Our method provides a proof of concept for the use of meta-cognition for language models as a strategy for more reliable and safer deployment.
We introduce ``meta-cognition'' as a term of art for machine learning systems design because we believe it will be an important component of future foundation model-based research.

Selective clarification QA involves interactive multi-turn dialogue.
Evaluating multi-turn dialogue with human participants is expensive, hard to reproduce, and can require ethics-board approval.
At the same time, automatic evaluations for multi-turn dialogue are often unreliable \citep{liu2016not, wahde2022conversational}.
In order to allow scalable model evaluation we describe an automatic prompt-driven evaluation protocol that allows a language model that is given privileged information to stand in for a person during evaluation.
In this way, we propose that language model research should shift towards \textbf{evaluation data-generating processes} rather than evaluation data sets.

In summary, our contributions are:
\begin{itemize}
    \item We introduce the \textbf{CLAM framework} for detecting ambiguous questions and clarifying them through multi-turn dialogue in LLMs (Section \ref{section_method}).
    \item We introduce the concept of language model \textbf{meta-cognition} as a general category of which CLAM is a proof-of-concept (Section \ref{s:meta-cognition}).
    \item We desribe an automatic evaluation protocol for \textbf{selective clarification QA} (Section \ref{section_automatic_evaluation_protocol}).
    \item We show that our implementation of CLAM delivers large accuracy improvements on data sets that contain ambiguous questions (Section \ref{section_experiments}).
\end{itemize}

In Section \ref{section_data_set}, we introduce our selective clarification QA task and the data set we create to evaluate it. 
In Section \ref{section_method}, we describe our framework for getting large language models to ask for clarification about ambiguous user requests. In Section \ref{section_automatic_evaluation_protocol}, we describe how we can automatically evaluate how good a language model is at asking for clarification about ambiguous user questions. In Section \ref{section_experiments}, we present our experiments and show that our method leads to large performance improvements in answering ambiguous questions while not affecting the performance on answering unambiguous questions. We review related work in Section \ref{section_related_work}. In Section \ref{section_conclusion}, we summarize our results and draw conclusions.

\section{Selective Clarification QA Data Set}
\label{section_data_set}

In this section, we introduce a data set that allows us to evaluate the performance on \textit{selective clarification QA} (\cref{figure_selective_clarification}). Selective clarification QA models the real-world observation that some user questions will be well-defined and thus will not need clarification, while other user questions will be ambiguous, and require clarification. The desired behaviour of a language model is to directly provide answers to unambiguous questions without asking for unnecessary clarification and on the other hand, ask the user for clarification if the user's question is ambiguous.

Existing data sets such as ClariQ \citep{aliannejadi2020convai3} and CLAQUA \citep{xu2019asking} do not allow us to evaluate models on all the steps of the selective clarification QA setting, see \cref{table_datasets_and_steps}. We describe existing data sets and their limitations in \cref{section_experiments}.

To study this setting, we therefore introduce a data set of pairs of questions that we call \textbf{Ambiguous TriviaQA}.
For each pair, there is one ambiguous question and one precisely disambiguated question.
We construct the data set so that just one piece of clarifying information is needed to make an ambiguous question precise.
In Section \ref{section_automatic_evaluation_protocol}, we explain how these sets of ambiguous and unambiguous questions let us automatically provide clarify ambiguous questions.

Our data set consists of 200 pairs of ambiguous and unambiguous questions that we derive from TriviaQA \citep{joshi2017triviaqa}. Given a randomly sampled TriviaQA question, we derive an ambiguous question by either:

\begin{itemize}
    \item Replacing a name or noun with a generic pronoun, e.g. ``Where in England was Dame Judi Dench born?'' becomes ``Where in England was she born?''.
    \item Replacing a noun phrase with a class the noun belongs to, e.g. ``Which country is Europe's largest silk producer?'' becomes ``Which country is Europe's largest producer?''
\end{itemize}

We use closed-book TriviaQA questions, that is, questions that stand alone and for which no accompanying context is provided. An additional advantage of deriving a data set from TriviaQA is that the reference answers are typically short, and only contain few non-essential words both of which increase the reliability of automatic accuracy metrics to evaluate models.

\section{CLAM: Selective clarification Framework}
\label{section_method}

In this section, we introduce \textbf{CLarify-if-AMbiguous (CLAM)}---a framework for language models to ask for selective clarification about possibly vague user questions.

The framework involves four stages, illustrated in Figure \ref{figure_selective_clarification}.
In the first stage, the user asks a language model a question.
The question is then classified as \texttt{ambiguous} or \texttt{not ambiguous}.
For questions that are not ambiguous, we return the answer immediately.
However, when questions are ambiguous, we generate a disambiguating follow-up question, as illustrated in \cref{figure_prompting_diagram}.
The model then uses the entire dialogue, including clarifying information, to answer the original question as it was intended.
Although we complete only a single iteration, it is also possible to recur the clarification process until the entire dialogue is considered to unambiguously ask a precise question which is an interesting direction for future research. 
We describe this formally in Algorithm \ref{alg:selective_clarification}.

Implementing the CLAM framework requires choosing a specific technique for:
\begin{itemize}
    \item classifying questions as ambiguous or not ambiguous;
    \item producing a clarifying question to follow up with.
\end{itemize}

In this paper, we implement both of these steps using prompting.
To classify a question's ambiguity, we prompt the model using few-shot prompts consisting of examples of ambiguous and unambiguous questions from the given data set. For Ambiguous TriviaQA, for instance, we use the following prompt:
\begin{quote}
Q: Who was the first woman to make a solo flight across this ocean? \\
This question is ambiguous: True.

Q: Who was the first woman to make a solo flight across the Atlantic? \\
This question is ambiguous: False. 

Q: In which city were Rotary Clubs set up in 1905? \\
This question is ambiguous: False.

Q: Who along with Philips developed the CD in the late 70s? \\
This question is ambiguous: False. 

Q: Where is the multinational corporation based?\\
This question is ambiguous: True.

Q: [question to be classified] \\
This question is ambiguous:
\end{quote}

We then take the model's log probability of the next token being \texttt{True} as a continuous predictor of whether the given question is ambiguous or not.

Interestingly, the fact that this works means that SotA models are able to detect ambiguous questions but do normally not ask the user for clarification.
We conjecture that this is the case because there are few dialogues including clarifying questions in the pre-training or finetuning data sets of these models. We leave a thorough investigation of why current models do not ask for clarification for future work.

We then produce a clarifying question using another prompt.
We simply append the string:

\begin{quote}
    ``In  order to answer this question, I have to ask the following clarifying question:''
\end{quote}

to the original ambiguous question.

We use a zero-shot prompt on the Ambiguous TriviaQA data set, and a few-shot prompt on the more challenging CLAQUA data sets. We describe the data set specific prompts in \cref{appendix_datasets_and_prompts}.

We then present the user with the clarifying question that is generated by the model based on this newly constructed prompt. 
Lastly, we present the combination of ambiguous question, clarifying question and the user's clarification to the language model to generate a final answer to the original question.
\subsection{Meta-cognition}
\label{s:meta-cognition}

In humans, meta-cognition is the process of thinking about your own thought process.
Working with foundation models, meta-cognition can involve using information that we have about the default sequence completion task to choose a new sequence completion task to perform instead.

In this paper, we use the example of prompting the model to detect whether a given question is ambiguous to then ask the user a clarifying question, rather than directly trying to answer the ambiguous question.
But in broader contexts, one can imagine many useful pipelines in which a secondary classifier is used to `redirect' the `thought process' of a foundation model.
For example, a classifier detecting some form of toxicity might be able to redirect the question to a more constructive frame by selecting an alternative prompt, allowing it to answer the question in a less toxic way.
Using this sort of pipeline, predictable failure modes can be gracefully and easily recovered from without resulting in any errors that are visible to the user. Chain-of-thought prompting \citep{wei2022chain} can be thought of as an implicit form of meta-cognition, while a more sophisticated pipeline might use additional information about the problem to shape the prompt construction.
\vspace{-1em}
\section{Automatic Evaluation Protocol}
\label{section_automatic_evaluation_protocol}
In this section, we introduce an automatic evaluation protocol that allows us to evaluate multi-turn dialogues without requiring human input. In the selective clarification QA setting (\cref{figure_selective_clarification}), the user provides clarification when the model asks a clarifying question about an ambiguous user input. We show that we can automate this step of providing clarification using a language model that is given privileged information about the ambiguous question. On a high-level, we thus suggest that model evaluation should move towards \textit{evaluation data-generating processes} rather than evaluation data sets.

The automated evaluation of dialogue systems is attractive given that human evaluations have numerous problems.
They are too expensive for most researchers to access.
They cannot be easily reproduced and are idiosyncratic in ways that are hard for external researchers to observe and critique (unlike automatic evaluations, whose flaws are relatively easy to examine).
Lastly, in many cases experiments involve humans can create additional ethics risks that in even the best cases incur additional administrative and ethical approvals costs, which can reduce research iteration speed, and in the worst cases create hazards for research participants.

Instead, we therefore use a language model to provide clarifying information when asked. This then allows us to automatically evaluate performance on the selective clarification task. Using machine learning models to simulate users to evaluate dialogue systems has been suggested before (see e.g. \citet{su2016continuously}). To the best of our knowledge, our work is the first to suggest that a parallel corpus of unambiguous and ambiguous questions can be used to prompt large language models to provide clarifying information about ambiguous questions.

\begin{figure}
    \centering
    \includegraphics[scale=0.2]{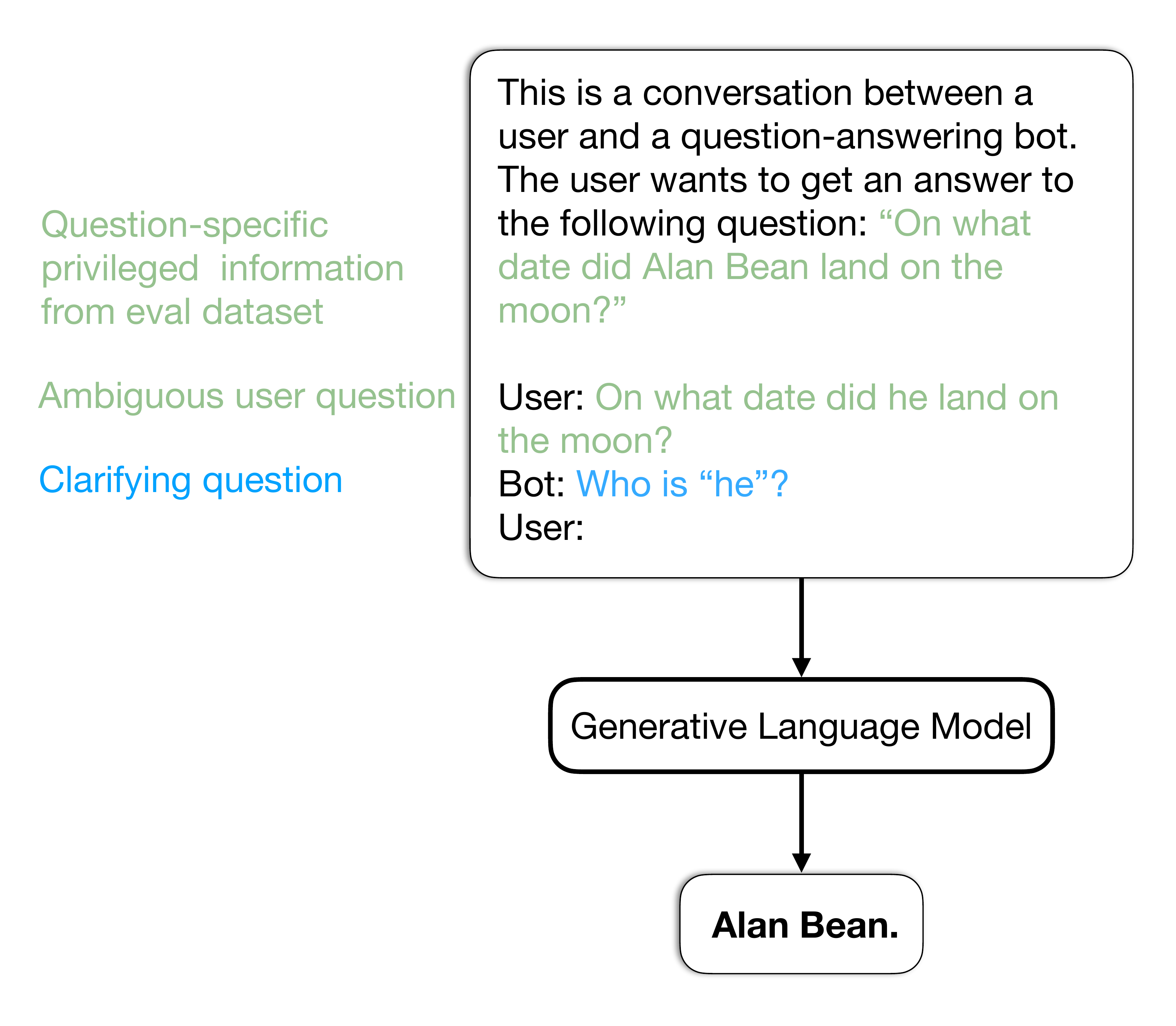}
    \caption{\textbf{Using a language model to provide clarification.} We prompt a language model to provide clarifying information given a clarifying question about an ambiguous user input. Our parallel corpus of ambiguous and corresponding unambiguous questions allows us to provide the unambiguous question to the oracle LM, based on which it can then provide appropriate clarifying information about the ambiguous question. See Figure \ref{figure_prompting_diagram} for a description of the other conversational turns.}
    \label{figure_oracle_prompt}
\end{figure}

For selective clarification QA, the language model may ask the user for clarifying information about the user's initial question (see Figure \ref{figure_prompting_diagram}).
Instead of a human, in our protocol, an `oracle' language model which has access to privileged information about the unambiguous question provides the clarifying information when asked (see Figure \ref{figure_oracle_prompt}).
Since our data set contains both ambiguous
and corresponding unambiguous questions, we provide the oracle model with privileged information by including the unambiguous question in its prompt. When appropriately asked for clarification, the oracle can then reliably provide clarifying information based on the unambiguous question in its prompt.

\vspace{-1em}

\section{Experiments}
\label{section_experiments}
In this section, we experimentally validate the ability of CLAM to successfully detect and seek clarification to ambiguous questions.

We first demonstrate its overall improved accuracy on the end-to-end task of question answering on a data set that contains both ambiguous and unambiguous questions. 

We then evaluate the performance on CLAM on each step of the pipeline (\cref{figure_selective_clarification}) that involves the language model individually: step 1b, distinguishing ambiguous from unambiguous questions; step 2, generating appropriate clarifying questions; and step 4, giving correct final answers given clarifying information. 

Lastly, we evaluate whether our automatic evaluation setup works reliably. That is, we test whether the oracle language model reliably provides the correct clarification when presented with a clarifying question.

\textbf{Models:} In principle, our framework should apply to any large language model architecture. In our experiments, we use the \texttt{text-davinci-002} and \texttt{davinci} models via the OpenAI API. \texttt{davinci} is a pre-trained language model similar to the original GPT-3 model \citep{brown2020language} and \texttt{text-davinci-002} \footnote{https://beta.openai.com/docs/model-index-for-researchers} is a GPT-3 model with an additional supervised fine-tuning stage on human-written demonstrations. These models are publicly available to researchers from any institution, which improves the reproducibility of our results and evaluation protocol. It can, however, be difficult to confirm whether the currently actively served version of the model is the same as the one used in these experiments, which is a challenge for reproducibility.

\textbf{Data sets:} There are many different types of ambiguity in text (see \citep{keyvan2022approach} for a taxonomy). To account for this, we evaluate CLAM on a range of data sets that cover different types of ambiguity. Our own data set \textit{Ambiguous TriviaQA} (described in \cref{section_data_set}) primarily covers under-specified user questions. \textit{ClariQ} \citep{aliannejadi2020convai3} is an information retrieval data set consisting of real search engine queries from the TREC Web Track \citep{clarke2012overview}. For each query human labellers also indicate on a scale of 1 to 4 to what extent clarification is needed. We convert ClariQ into a binary classification problem by labelling queries with \textit{clarification needs} of 1 and 2 as \texttt{unambiguous} and queries with \textit{clarification needs} of 3 and 4 as \texttt{ambiguous}. Furthermore, we evaluate CLAM on \textit{CLAQUA} \citep{xu2019asking}, a data set consisting of two sub-datasets that cover different types of ambiguity: a task that we call \textit{CLAQUA I}, in which a proper noun in the question can refer to different entities, and a task that we call \textit{CLAQUA II} which consists of multi-turn conversation where the last turn is a question that could refer to multiple different previous conversational turns. To reduce the costs of our experiments from calling the OpenAI API, we use random sub-samples of 400 of each data set when evaluating ambiguity detection, and 100 random samples from each data set to evaluate the generation of clarifying questions and question answering accuracy.

However, only our data set can be used to evaluate all of the steps of our pipeline end-to-end.
Because of limitations in the design of the other data sets, which were designed for different purposes, they can only be used for a subset of the steps in our pipeline.
\cref{table_datasets_and_steps} provides an overview of which data set can be used to test which of the steps.

\begin{table}[t]
\caption{Overview of which data sets can be used to evaluate which of the steps of selective clarification QA.}
\label{table_datasets_and_steps}
\vspace{-3mm}
\begin{center}
\begin{small}
\begin{sc}
\resizebox{\linewidth}{!}{
\begin{tabular}{lcccc}
\toprule
& ClariQ & CLAQUA I & CLAQUA II  & \makecell{Ambiguous \\ TriviaQA (Ours)} \\
\midrule
\makecell[l]{1b: Detect ambiguity}  & $\surd$ & $\surd$ & $\surd$ & $\surd$\\
\makecell[l]{2: Ask clarifying\\\quad questions}& & $\surd$& $\surd$ & $\surd$  \\
\makecell[l]{2-4: Final accuracy\\\quad(ambiguous only)}& & $\surd$& $\surd$ & $\surd$  \\
\makecell[l]{2-4: Final accuracy\\\quad(unambiguous only)}& & &  & $\surd$  \\
\bottomrule
\end{tabular}
}
\end{sc}
\end{small}
\end{center}
\vskip -0.1in
\end{table}

\textbf{Metrics: }We measure the accuracy of a model answer by evaluating whether the model answer contains the reference answer. This accounts for the fact that the language model often answers in full sentences while the reference answer consists only of the target terms themselves.

One challenge in the automatic evaluation of free-form generations is that one answer can be expressed in many different ways, see e.g. \citet{kuhnsemantic}. To account for this, we manually evaluate the accuracy metric on all of our data sets. Similarly, we manually evaluate whether the language models ask appropriate clarifying questions.

In addition to the raw accuracy, we introduce an \textit{adjusted accuracy} which is suitable for selective clarification QA specifically. This measure penalizes the language model system for asking unnecessary clarifying questions about unambiguous questions.
To adjust the accuracies we begin with a score of 1 for a correct answer and 0 for an incorrect answer (as normal) but then multiply it with 0.8 if the question is unambiguous and the model nonetheless asks for clarification.
The specific value of 0.8 is arbitrary and our results hold for a range of penalty terms, see Table \ref{table_adjusted_accuracy_results}.

To evaluate how well the different methods can distinguish ambiguous from unambiguous questions, we measure the commonly used Area Under the Receiver Operator Characteristic Curve (AUROC). The AUROC metric is equivalent to the probability that a randomly chosen correct answer has a predictor score than a randomly chosen incorrect answer. Higher scores are better, with perfect uncertainty scoring 1 while a random uncertainty measure would score 0.5.

\textbf{Baselines: } We compare CLAM to three baselines. \textit{Default GPT} prepends the question with a question-answering prompt: \texttt{This is a conversation between a user and a question-answering bot}.

\textit{Prompting baseline} uses a prompt that explicitly instructs the model to selectively ask the user for clarifying questions: \texttt{This is a conversation between a user and a question-answering bot. The bot asks the user for clarification if the user's question is ambiguous or imprecise.}

\textit{Force clarification} does not distinguish between ambiguous and unambiguous questions, and always prompts the language model to ask for clarification about the user's input.

\begin{figure}[h]
    \centering
    \centering
    \begin{subfigure}[b]{0.45\textwidth}
        \centering
        \includegraphics[width=0.8\textwidth]{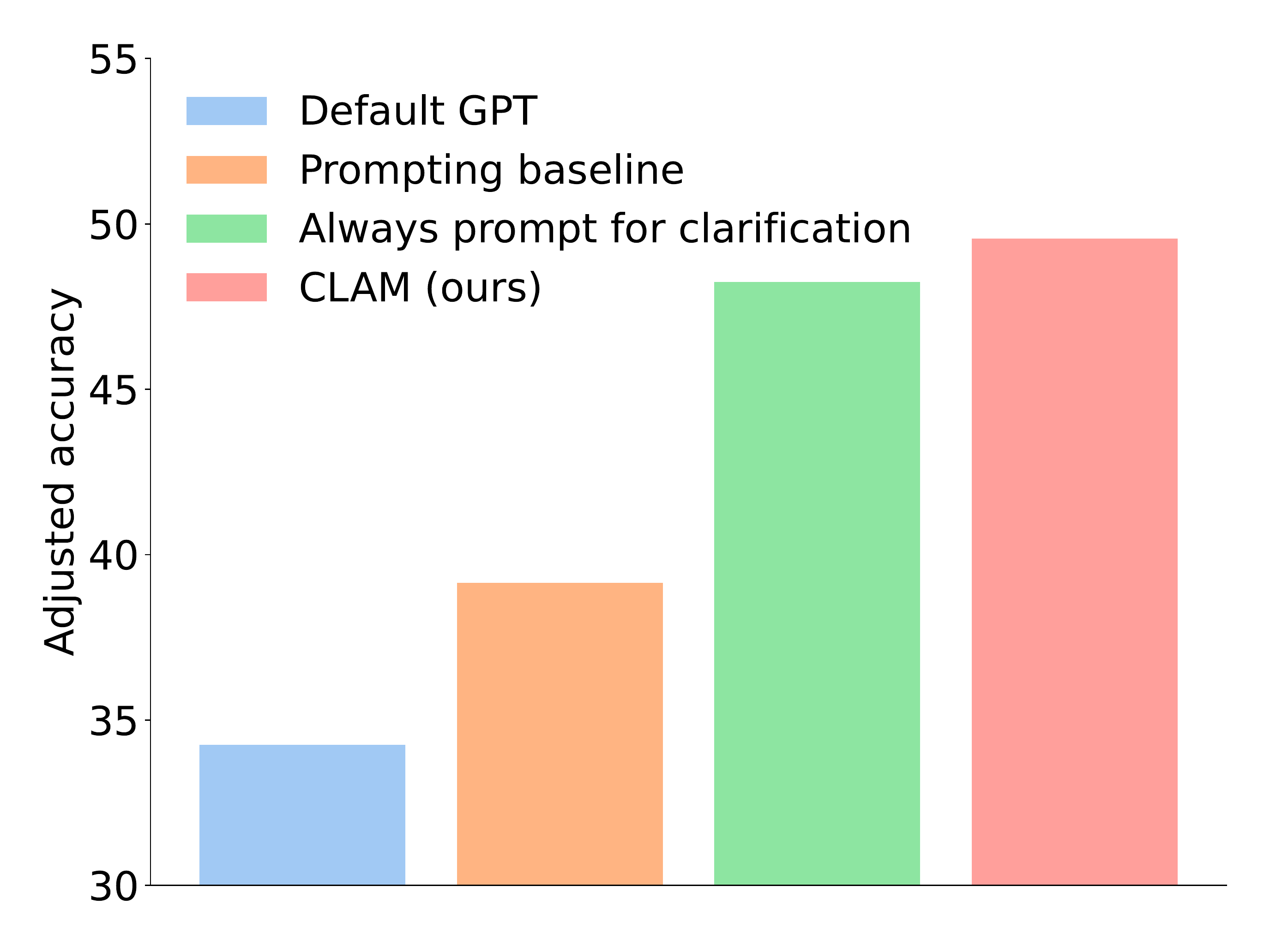}
        \vspace{-4mm}
        \caption{}
        \label{figure_main_result}
    \end{subfigure}
    \begin{subfigure}[b]{0.45\textwidth}
        \centering
        \includegraphics[width=0.8\textwidth]{
        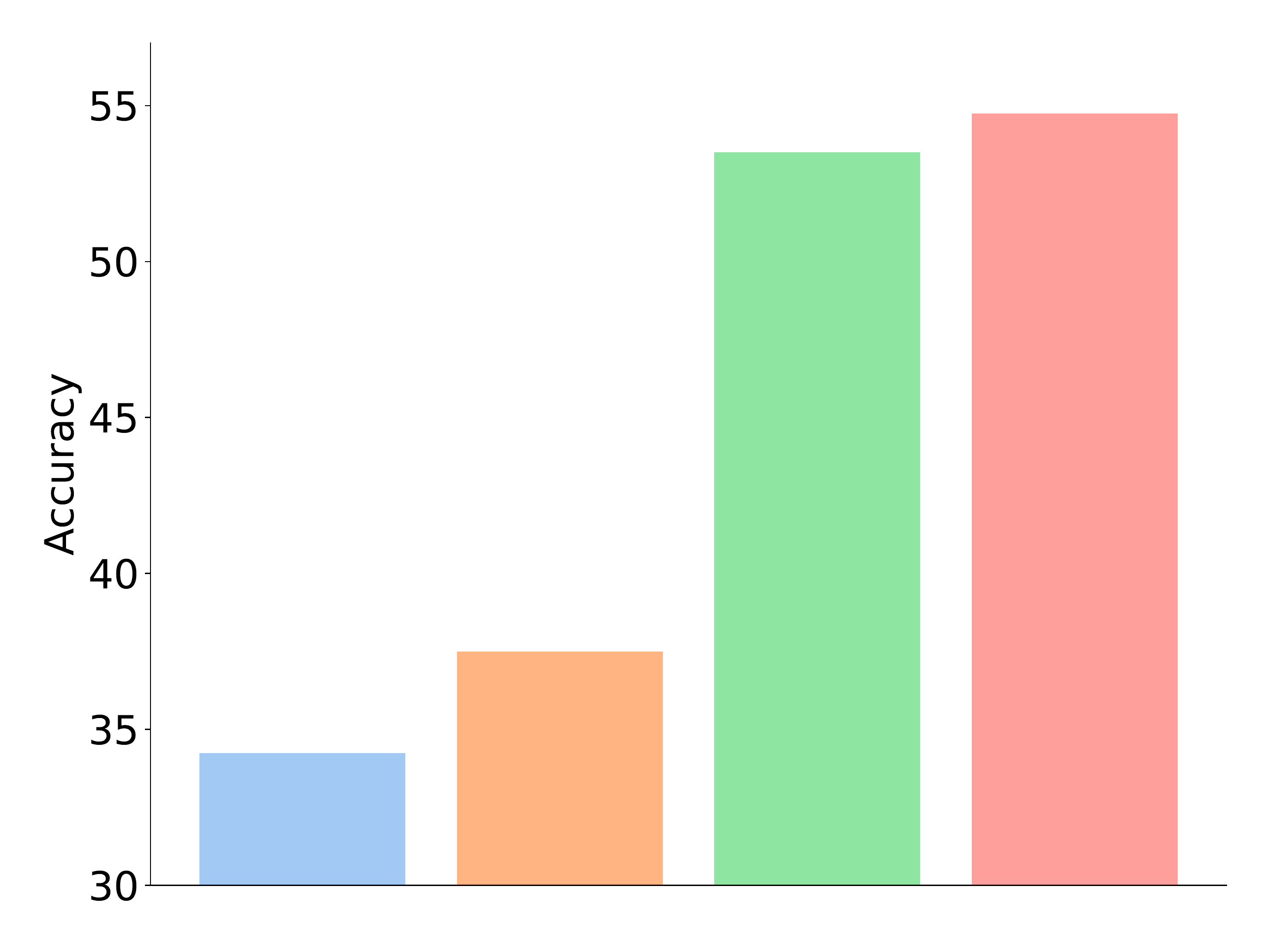}
        \vspace{-4mm}
        \caption{}
        \label{fig:mixed}
    \end{subfigure}
    \caption{ \textbf{CLAM improves question-answering accuracy on a set of ambiguous and unambiguous Trivia questions}. (a) CLAM clarifies ambiguous questions without asking for unnecessary clarification on unambiguous questions (which is reflected in the adjusted accuracy metric). Always prompting the language model to ask the user for clarification increases the accuracy on ambiguous questions but incurs a penalty on unambiguous questions. The prompting baseline rarely asks the user for clarification and thus only improves the accuracy slightly. (b) \textbf{Accuracy on full data set:} Without penalizing unnecessary clarifying questions, always prompting for clarification and CLAM perform comparably well, and much better than default GPT and the prompting baseline. CLAQUA and ClariQ only cover parts of the selective clarification pipeline which is why only TriviaQA results are reported here.}
\end{figure}

\subsection{Results}

We first establish the overall accuracy improvement of our implementation of the CLAM framework on selective clarification QA (steps 1 through 4 in \cref{figure_selective_clarification}). Then, we study the performance on each of the necessary steps in \cref{figure_selective_clarification}: detecting ambiguous questions (step 1b), generating useful clarifying questions (step 2), and giving final answers after receiving clarification (step 4). Lastly, we show that our automatic evaluation scheme for selective clarification QA works reliably, by showing that the oracle language model correctly provides clarification when asked (step 3). 

We report the results of the \texttt{text-davinci-002} model in the main part of the model and provide the results for \texttt{davinci} in the Appendix. \texttt{davinci} performs less well across all tasks but the differences between the different methods are qualitatively similar to those of the \texttt{text-davinci-002} model.

ClariQ and CLAQUA each only contain components to evaluate some of the parts of the pipeline (see \cref{table_datasets_and_steps}), and some figures thus report results on a subset of the data sets.

\subsubsection{Overall performance}

Overall, we find that CLAM boosts the language model's adjusted accuracy on ambiguous TriviaQA (containing both ambiguous and unambiguous questions) by roughly 20 percentage points (Figure \ref{figure_main_result}). Recall that the adjusted accuracy multiplies the accuracy on a given question with a penalty term $\lambda = 0.8$ each time the model unnecessarily asks for clarification on unambiguous questions as described in Section \ref{section_experiments}. \textit{Force clarification} improves the accuracy on ambiguous questions but incurs a large penalty on unambiguous questions for unnecessarily asking for clarification. Using the \textit{Prompting baseline} only leads to a moderate accuracy improvement. Without the penalty term for asking unnecessary clarifying questions, the performance of always prompting for clarification and \textit{CLAM} on the data set of both ambiguous and unambiguous questions is comparable, and they both clearly outperform default GPT and the prompting baseline (\cref{fig:mixed}).

On the ambiguous questions only, we find that, as expected, both \textit{Force clarification} and \textit{CLAM} greatly improve the question-answering accuracy as compared to the default GPT performance, see appendix Figure \ref{fig:ambig_only}. Importantly, always clarifying and CLAM almost entirely close the gap on performance on ambiguous questions as compared to the performance on unambiguous questions, see appendix Figure \ref{fig:unambiguous_only}. On the unambiguous questions, the different methods generally do not affect the model performance as compared to the default GPT behaviour. However, always prompting for clarification always leads to unnecessary turns of conversation on unambiguous questions which is bad for the user experience, and sometimes actually hurts the accuracy by leading the conversation off-topic.

\begin{figure}
    \centering
    \includegraphics[scale=0.25]{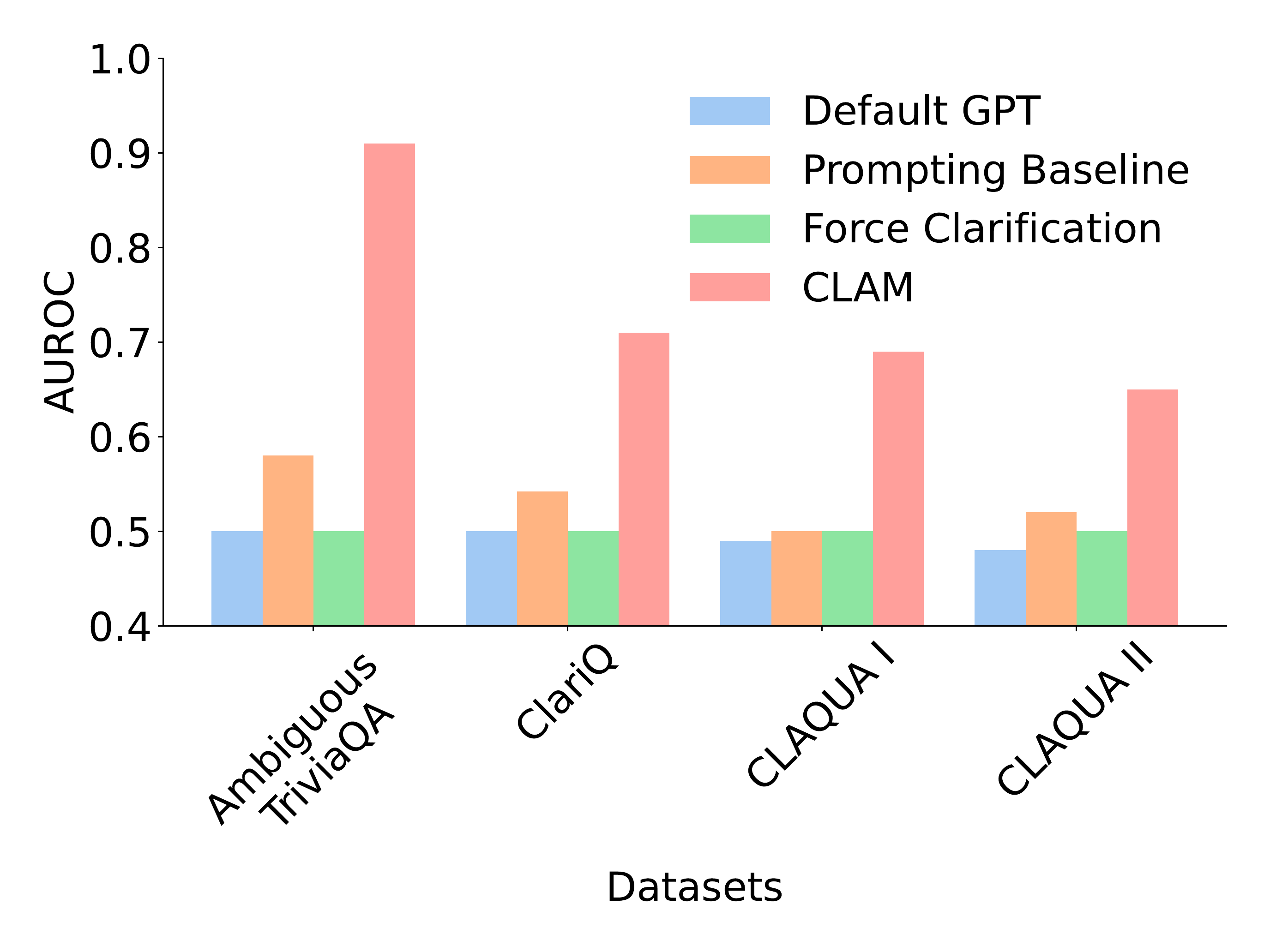}
    \caption{\textbf{CLAM reliably distinguishes ambiguous from unambiguous questions}. The AUROC measures how accurately the different methods predict whether a given question is ambiguous or not. \textit{Default GPT} almost never asks for clarification whereas \textit{Force clarification} always asks for clarification. Both methods thus do not distinguish ambiguous from unambiguous questions. }
    \label{figure_ambiguity_detection}
\end{figure}
\subsubsection{Individual pipeline components}

\textbf{Step 1b: Detect Ambiguity}

\textit{CLAM} reliably distinguishes ambiguous from unambiguous questions, see \cref{figure_ambiguity_detection}. The few-shot prompting-based log probability of a given question being ambiguous or unambiguous achieves high AUROCs on all data sets.
\textit{Default GPT} almost never asks for clarification and thus achieves an AUROC of roughly 0.5 on all data sets. \textit{Force clarification} on the other hand treats all questions as ambiguous questions, and thus always has an AUROC of 0.5. The \textit{Prompting baseline} does sometimes asks for clarification on ambiguous questions and almost never asks for clarification on unambiguous questions, and is thus slightly better than random at detecting ambiguity.

\textbf{Step 2: Ask clarifying questions}

We manually label 100 randomly selected pairs of ambiguous questions and model-generated corresponding clarifying questions for each data set to test how reliably the language model generates the correct clarifying question (Table \ref{table_human_eval}). According to our judgment, the model generates the correct clarifying question for 84\%, 99\% and 95\% of the questions in Ambiguous TriviaQA, CLAQUA I and CLAQUA II respectively. We find that on TriviaQA the model sometimes asks incorrect clarifying questions either by just repeating the given ambiguous question or by asking an unrelated additional question about the subject of the ambiguous question.

\begin{table}
\caption{\textbf{LMs can accurately generate clarifying questions and provide clarification during evaluation}. Human evaluation of each of the conversational turns. \textit{ClariQ} does not contain answers for the ambiguous questions and is thus not displayed in this table.}
\label{table_human_eval}
\vskip 0.15in
\begin{center}
\begin{small}
\begin{sc}
\resizebox{\linewidth}{!}{
\begin{tabular}{lcccr}
\toprule
Method & \makecell{Ambiguous \\TriviaQA} & CLAQUA I & CLAQUA II \\
\midrule
    \makecell[l]{Step 2:Correct \\\quad clarifying question} & 84.0\%& 99.0\%& 95.0\%\\
\makecell[l]{Step 3: Correct \\\quad oracle answer}  & 98.8\%& 67.7\%  & 98.7\% \\

\bottomrule
\end{tabular}
}
\end{sc}
\end{small}
\end{center}
\vskip -0.1in
\end{table}

\textbf{Steps 2 through 4: Final answer after clarification}

\begin{figure}
    \centering
    \includegraphics[scale=0.25]{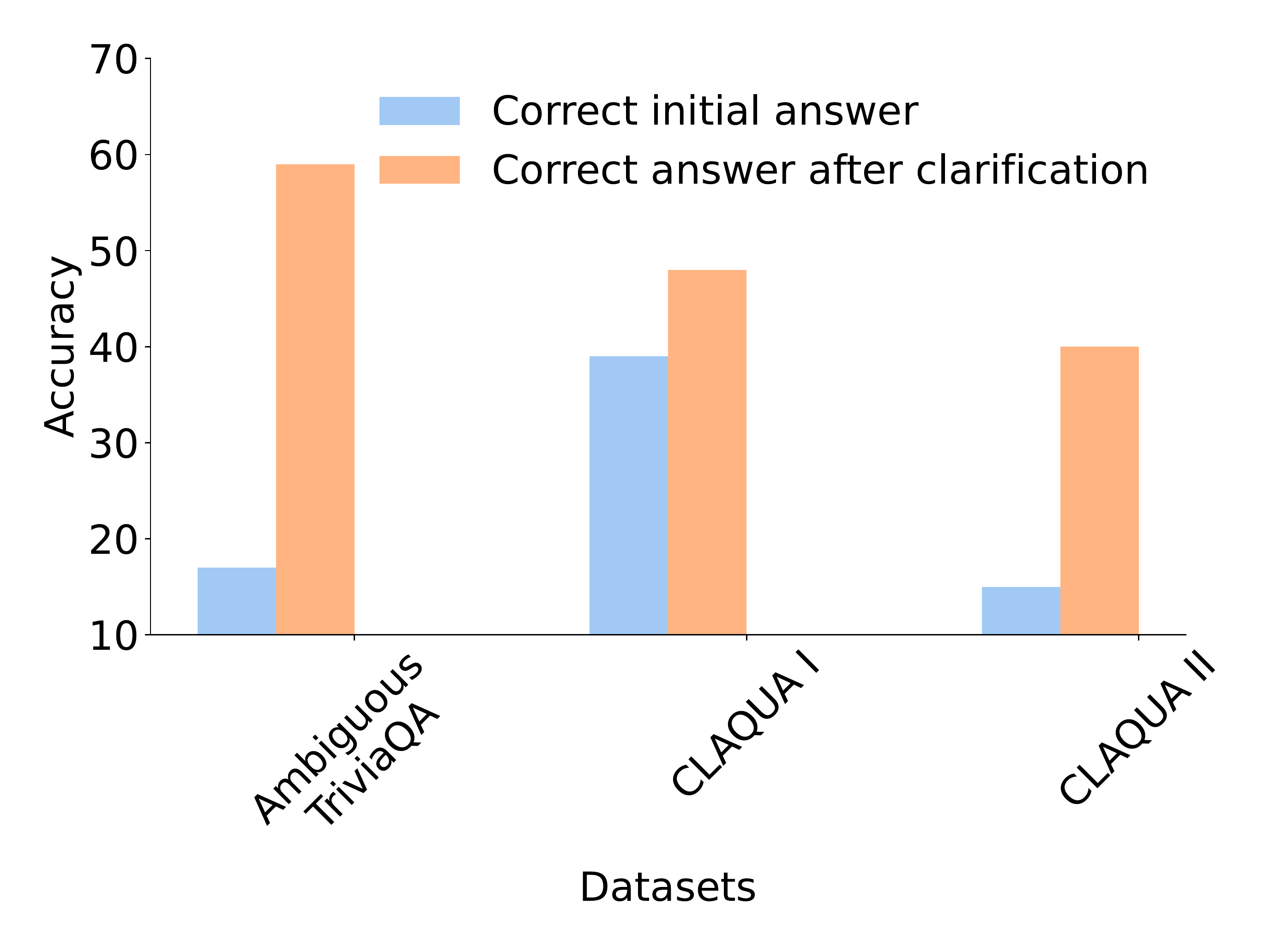}
    \caption{\textbf{Prompting the model to ask the user for clarification improves QA accuracy on ambiguous questions.} ClariQ does not contain answers for the ambiguous questions and is thus not displayed in this figure.}
    \label{figure_accuracy_improvement_on_ambiguous_questions}
\end{figure}

Ignoring the task of detecting ambiguity and focusing only on questions which are known to be ambiguous, asking the user for clarification lets the model answer ambiguous questions much more accurately for all data sets (\cref{figure_accuracy_improvement_on_ambiguous_questions}).

\textbf{Step 3: Oracle automatic evaluation}

Our oracle language model (\cref{section_automatic_evaluation_protocol}) provides accurate clarifying information given a clarifying question by the language model under evaluation. We manually label 100 randomly sampled conversations based on ambiguous questions from Ambiguous TriviaQA, CLAQUA I and CLAQUA II. We report for what fraction of correct clarifications the oracle language model provides.

The oracle model reliablly provides the correct clarifying information on TriviaQA and CLAQUA II, see Table \ref{table_human_eval}. On CLAQUA I the clarifying information has to disambiguate two possible meanings of a proper noun. We find that, while the model is generally able to do so, it will sometimes provide information that applies to both possible entities, and thus does not correctly disambiguate the two options.

\section{Related work}
\label{section_related_work}

A range of approaches for i) detecting ambiguous questions and then ii) asking clarifying questions have been proposed  in information retrieval and in the conversational search literature (see \citet{keyvan2022approach} for an overview). In terms of \textit{detecting} ambiguous queries, \citet{trienes2019identifying} use a logistic regression to identify ambiguous queries based on the characteristics of similar queries. \citet{dhole2020resolving} use a BiLSTM model to distinguish ambiguous from unambiguous queries. In terms of \textit{generating clarifying questions}, both rule-based and neural network-based approaches have been proposed. \citet{wang2021template}, for instance, use a clarifying question template that is completed with words from a vocabulary. \citet{dhole2020resolving} frame disambiguation as distinguishing between different plausible user intents for a given question. They use a set of syntactic transformations of a given ambiguous question, and then select a clarifying question that will best disambiguate between different user intents. \citet{rao2019answer} use a GAN to generate clarifying questions.

AmbigQA \citep{min2020ambigqa} is an alternative data set intended to explore ambiguous question answering derived from Natural Questions \citep{kwiatkowski_2019_natural_questions}. However, \citet{min2020ambigqa} note that the ambiguity in their data set is ``sometimes subtle'' and that ``many [ambiguities] are only apparent after examining one or more Wikipedia pages''. Furthermore, \citet{krasheninnikov2022assistance} then find that the performance of AmbigNQ, a baseline they introduce to find various precise questions for a given ambiguous question, is low, attesting to the difficulty of this task. Given how difficult AmbigNQ seems to be, we suggest that our Ambiguous TriviaQA data set, which requires less factual knowledge, is a more effective evaluation tool for selective clarification QA.

In transformer-based dialogue systems, however, dealing with ambiguous queries has received little attention so far. To the best of our knowledge, \citet{krasheninnikov2022assistance} (concurrent with our work), is the only paper that addresses ambiguous question resolution in GPT-like language models. The authors fine-tune a 175B parameter GPT-3 model on a data set of conversations consisting of ambiguous user requests, clarifying questions, and final answers. They show that fine-tuning the model on this data set leads to a slight accuracy improvement in answering ambiguous questions derived from AmbigQA \citep{min2020ambigqa}. The authors note that under this approach the model often does not recognize ambiguous inputs (false omission rate of 44.5\%). We further note that in contrast to this method our approach does not require any fine-tuning, neither to improve the performance of the question-answering nor for the oracle model. 

\section{Conclusion}
\label{section_conclusion}
In this paper, we introduce CLAM, a framework for selective clarification QA with large language models which detects and resolves ambiguity by asking clarifying questions.
We provide this as an example of meta-cognition in foundation models.
We implement the framework using few-shot prompting.
This additional clarifying conversational turn significantly increases language models' question-answering accuracy on ambiguous questions without affecting unambiguous questions.
Moreover, we show that few-shot prompting is a highly reliable way of detecting whether a given question is ambiguous which allows us to answer clarifying questions about ambiguous questions only and avoid asking unnecessary questions about precise user inputs.

In order to support scalable research, we motivate a shift towards evaluation data-generating processes and introduce a method to automatically evaluate multi-turn dialogues involving ambiguous questions using an oracle language model with access to extra information.
\newpage

\section*{Acknowledgements}

Sebastian Farquhar carried out this work in his capacity as an associate member of the OATML lab, but he is also employed by DeepMind.

We are grateful to Geoffrey Irving and Laura Rimell for their advice and feedback on earlier drafts of this paper. We are also grateful to the members of the OATML group for helpful discussions about this project.

\bibliography{example_paper}
\bibliographystyle{icml2022}

\newpage
\appendix
\onecolumn

\section{Additional experimental results}

\subsection{Results on \texttt{davinci} model}

\begin{figure}
    \centering
    \includegraphics[scale=0.3]{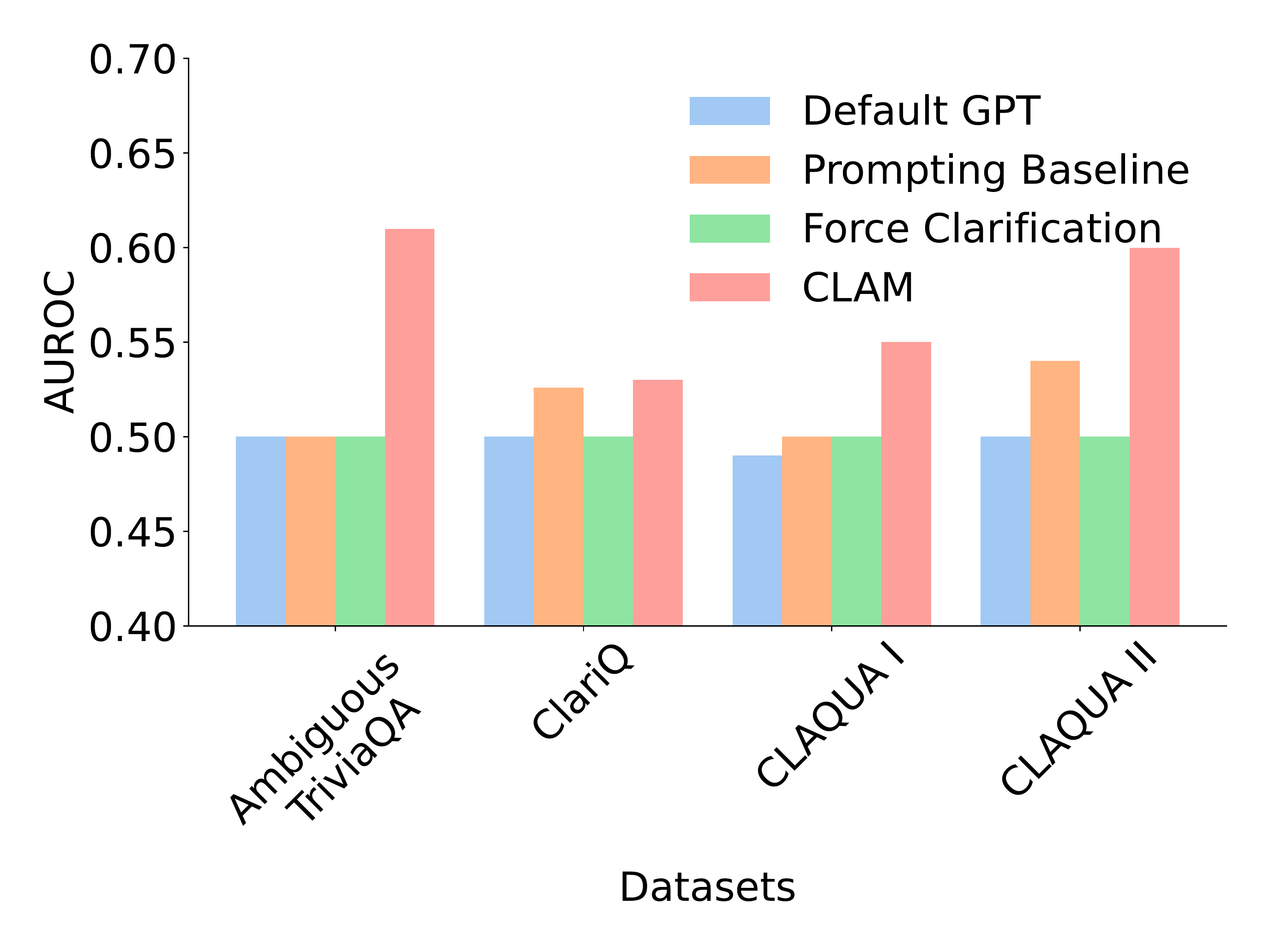}
    \caption{Using the \texttt{davinci} model for CLAM, we are able to detect ambiguous questions more reliably than \textit{Default GPT}, \textit{Force clarification} and \textit{Prompting baseline}. \texttt{davinci} detects ambiguous questions less well than the\texttt{text-davinci-002} model used in the main part of the paper.}
    \label{figure_ambiguity_auroc_davinci}
\end{figure}

\begin{figure}
    \centering
    \includegraphics[scale=0.3]{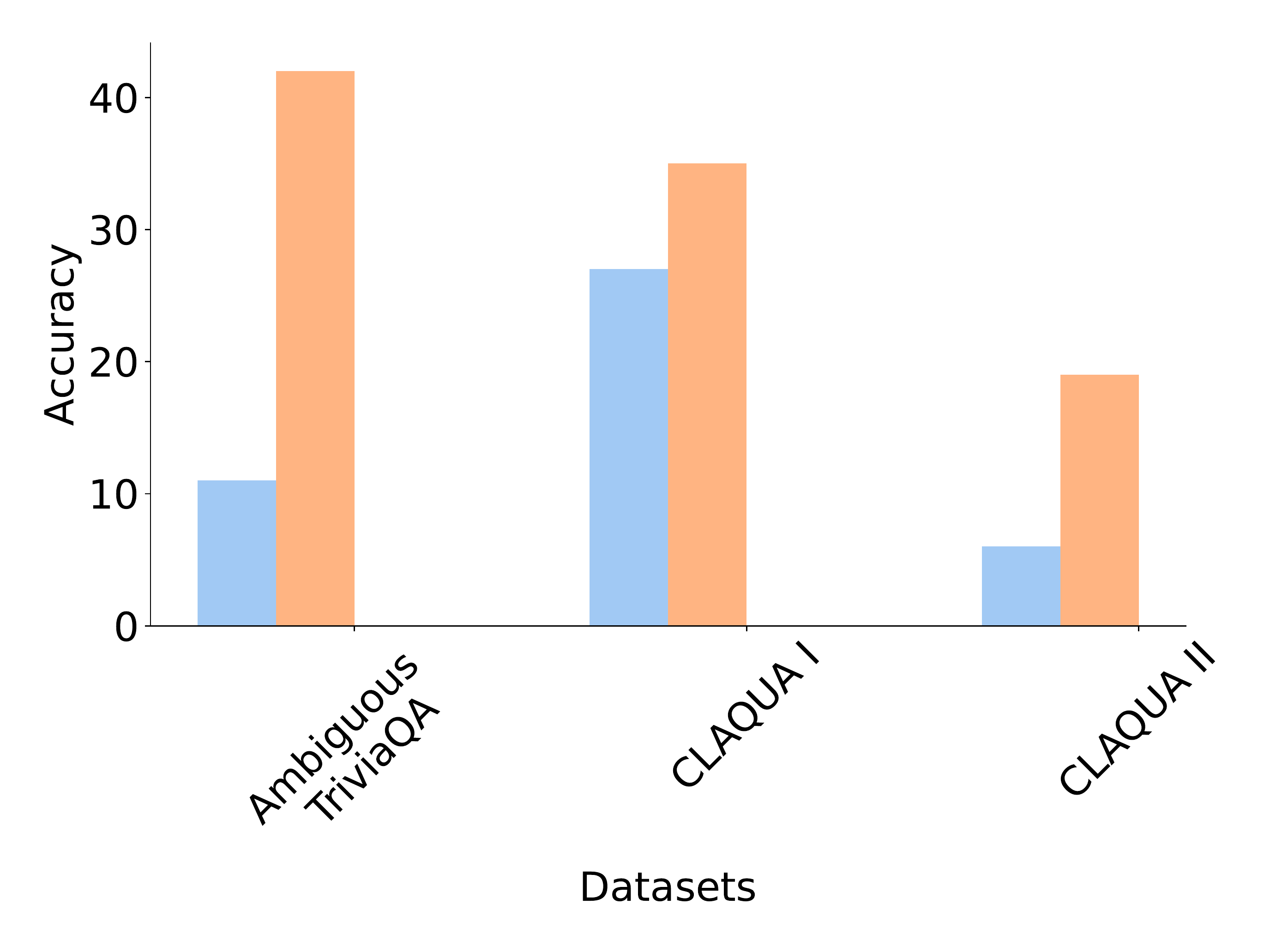}
    \caption{\textbf{\texttt{davinci}: Prompting the model to ask the user for clarification improves QA accuracy on ambiguous questions.} ClariQ does not contain answers for the ambiguous questions and is thus not displayed in this figure.}
    \label{figure_accuracy_improvement_on_ambiguous_questions_davinci}
\end{figure}

In this section, we report the results of our experiments using the \texttt{davinci} model whereas we used the \texttt{text-davinci-002} model in the main part of the paper. Overall, we observe qualitatively similar results using the \texttt{davinci} model as with the \texttt{text-davinci-002}, although at a lower level. 
In summary, we find that
\begin{itemize}
    \item We find that \texttt{davinci} is also able to distinguish ambiguous from unambiguous questions, see \cref{figure_ambiguity_auroc_davinci}, although less reliably than \texttt{text-davinci-002}.
    \item \texttt{davinci} is able to reliably generate clarifying questions across all data sets, and relatively reliably provides clarification, see \cref{table_davinci_clarification}.
    \item Lastly, we show that clarifying improves QA accuracy on ambiguous questions across all data sets \cref{figure_accuracy_improvement_on_ambiguous_questions_davinci}.
\end{itemize}

\begin{table}[t]
\caption{The \texttt{davinci} model reliably generates the correct clarifying question and clarification}
\label{table_davinci_clarification}
\label{sample-table}
\vskip 0.15in
\begin{center}
\begin{small}
\begin{sc}
\resizebox{0.7\linewidth}{!}{
\begin{tabular}{lcccr}
\toprule
 & Ambiguous TriviaQA & CLAQUA I & CLAQUA II \\
Correct clarifying question & 97\% & 99\% & 92\%\\
Correct oracle answer & 98\% & 76\% & 67\%  \\

\bottomrule
\end{tabular}
}
\end{sc}
\end{small}
\end{center}
\vskip -0.1in
\end{table}

\subsection{Ablation of penalty term in adjusted accuracy}
In \cref{table_adjusted_accuracy_results}, we show that our adjusted accuracy results hold up across different penalty terms $\lambda$. The results in the main part of the paper use $\lambda = 0.8$

\begin{table}
    \centering
    \caption{\textbf{Adjusted accuracy results for different penalty terms on the full Ambiguous TriviaQA data set.} In our adjusted accuracy metric, the accuracy of the subset of questions that are \textit{unambiguous} and on which the language model nonetheless asks for clarification are multiplied with a penalty term $0 < \lambda < 1$. We show that CLAM outperforms the default GPT model and the baselines regardless of the particular choice of $\lambda$. Note that default GPT and the prompting baseline never ask for clarification on unambiguous questions and thus do not incur a penalty.}
    \begin{tabular}{lcccccc}
        \toprule
        Method / $\lambda$ &0.5 &0.6  &0.7   &0.8   &0.9   &1.0
        \\ \midrule
        Default GPT & 34.25 & 34.25 & 34.25 & 34.25 & 34.25 & 34.25 \\
        Prompting baseline  & 37.50 & 37.50 & 37.50 & 37.50 & 37.50 & 37.50 \\
        Always prompt for clarification & 40.37 & 43.0 & 45.62& 48.25 & 50.88 & 53.5 \\
        CLAM (ours) & 53.88 & 54.05 & 54.22 & 54.40 & 54.58 & 54.75 \\

        \bottomrule
\end{tabular}
    \label{table_adjusted_accuracy_results}
\end{table}

\subsection{Additional analysis of CLAM performance}

\begin{figure}
    \centering
    \begin{subfigure}[b]{0.45\textwidth}
        \centering
        \includegraphics[width=0.9\textwidth]{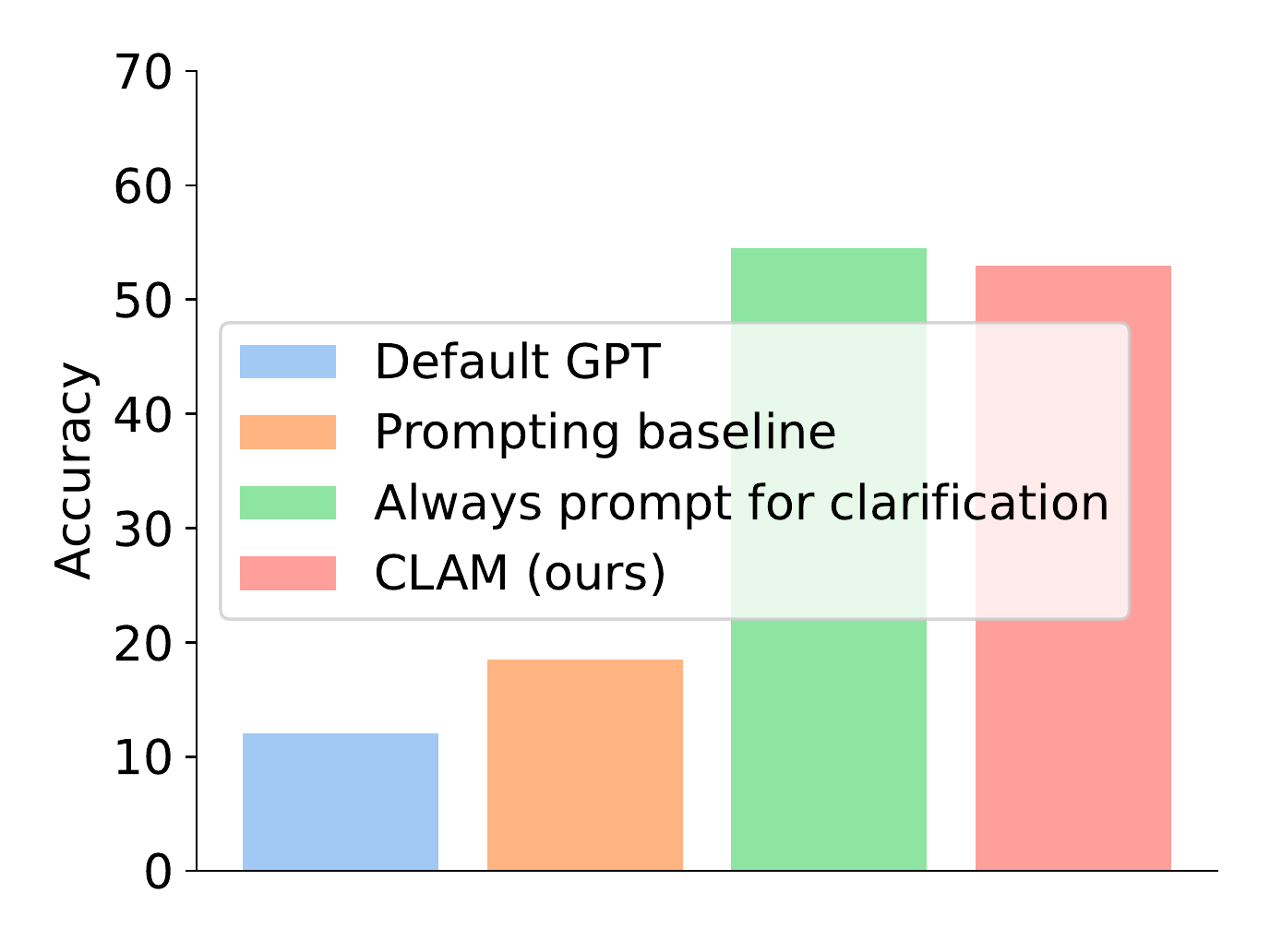}
        \vspace{-4mm}
        \caption{}
        \label{fig:ambig_only}
    \end{subfigure}
    \begin{subfigure}[b]{0.45\textwidth}
        \centering
        \includegraphics[width=0.9\textwidth]{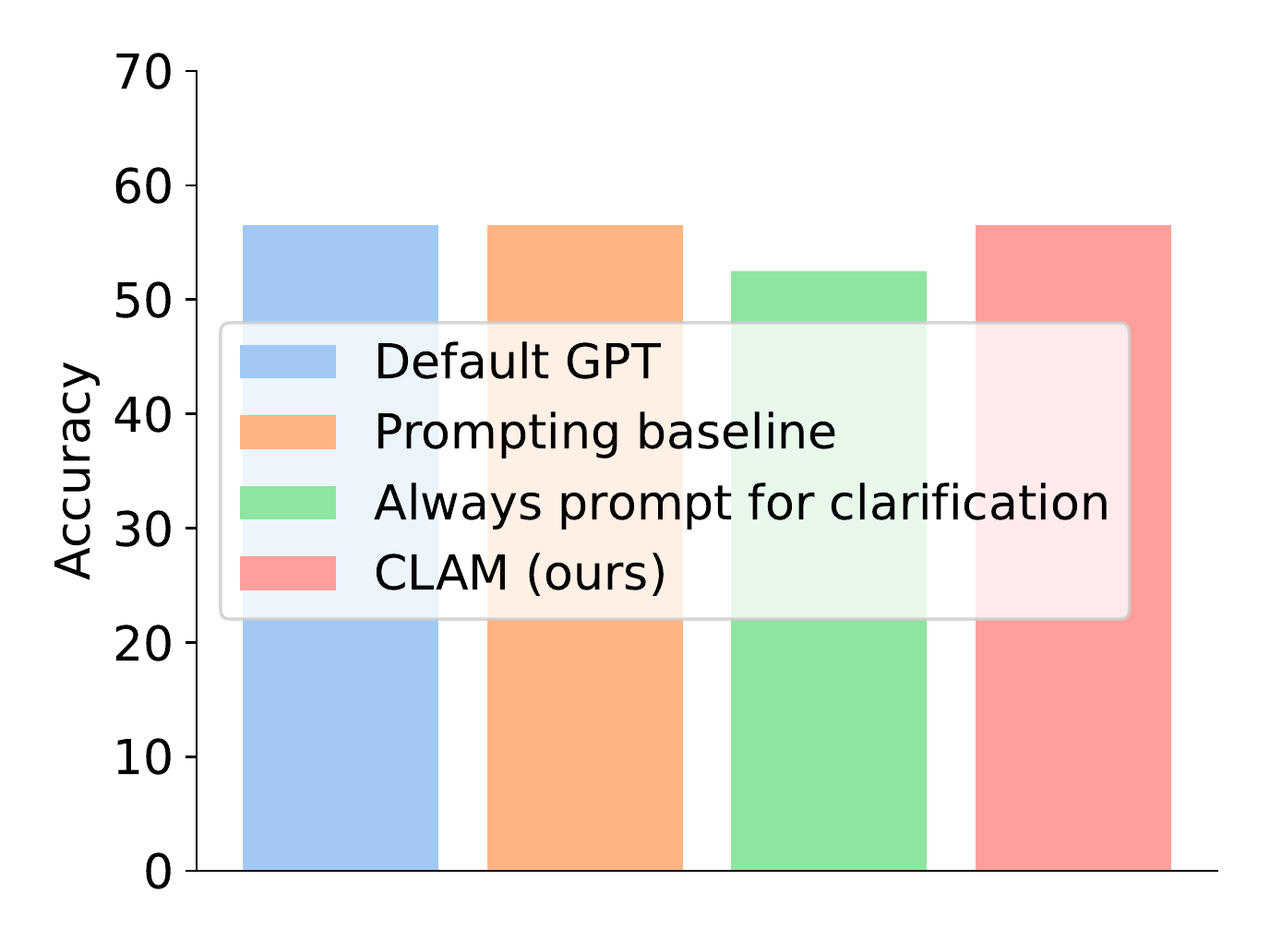}
        \vspace{-4mm}
        \caption{}
        \label{fig:unambiguous_only}
    \end{subfigure}
    \caption{(a) \textbf{Accuracy on ambiguous questions from Ambiguous TriviaQA only:} CLAM and always prompting the language model to ask the user for clarification yields large improvements in accuracy over the default GPT behaviour and the prompting baseline. (b) \textbf{Accuracy on unambiguous questions from Ambiguous TriviaQA only:} The model performance on unnecessary questions remains largely unaffected. Note that always prompting for clarification leads to unnecessary turns of conversation on unambiguous questions which is in itself undesirable and sometimes decreases the accuracy by leading the conversation off-topic.}
    \label{figure_trivia_qa_additional_results}
\end{figure}

In \cref{figure_main_result} in the main part of the paper, we show that CLAM improves the question-answering accuracy on the full Ambiguous TriviaQA data set. In this section, we additionally compare the different baselines on the ambiguous and unambiguous questions separately, see \cref{figure_trivia_qa_additional_results}. We find that, as expected, both \textit{Force clarification} and \textit{CLAM} lead to a large accuracy improvmenet on ambiguous questions, and largely leave the performance on unambiguous questions unaffected. 

\newpage
\section{Additional information on data sets and prompts}
\label{appendix_datasets_and_prompts}

In this section, we provide additional details on the data sets and prompts used in our experiments.

\subsection{Ambiguous TriviaQA} 

As described in \cref{section_data_set}, we derive the ambiguous TriviaQA from the original TriviaQA \citet{joshi2017triviaqa} data set. Given an unambiguous question from the TriviaQA data set, we derive an ambiguous question as follows:

\begin{itemize}
    \item Replacing a name or noun with a generic pronoun, e.g. ``Where in England was Dame Judi Dench born?'' becomes ``Where in England was she born?''.
    \item Replacing a noun phrase with a class the noun belongs to, e.g. ``Which country is Europe's largest silk producer?'' becomes ``Which country is Europe's largest producer?''
\end{itemize}

\textbf{Ambiguity detection}

\begin{verbatim}
Q: Who was the first woman to make a solo flight across this ocean? 
This question is ambiguous: True.

Q: Who was the first woman to make a solo flight across the Atlantic? 
This question is ambiguous: False. 

Q: In which city were Rotary Clubs set up in 1905? 
This question is ambiguous: False.

Q: Who along with Philips developed the CD in the late 70s? 
This question is ambiguous: False. 

Q: Where is the multinational corporation based?
This question is ambiguous: True.

Q: [question to be classified] 
This question is ambiguous:
\end{verbatim}

As explained in the main part of the paper, we then take the log probability of the next token being \texttt{True} as a continuous predictor of whether the given question is ambiguous or not.

\textbf{Clarifying question generation}

\begin{verbatim}
This is a conversation between a user and a question-answering bot.
User: On what date did he land on the moon?
Bot: To answer this question, I need to ask the following clarifying question:
Who is he?

###

User: Which country on this continent has the largest population?
Bot: To answer this question, I need to ask the following clarifying question:
Which continent? 

###

User: {initial_question}
Bot: To answer this question, I need to ask the following clarifying question:
\end{verbatim}

\subsection{ClariQ}

ClariQ \citep{aliannejadi2020convai3} is based on search engine queries from the TREC Web Track 2009-2012 data set \citep{clarke2012overview}. Every query in the data set has a human label \textit{clarification need} (1-4) that indicates how unclear the human labeller perceives this question to be. We convert this data into a binary classification data set by labelling queries with clarification needs of 1 and 2 as \textit{unambiguous}, and queries with clarification needs of 3 and 4 as \textit{ambiguous}.

This data set does not contain reference answers for the given questions. We can thus only use this data set to measure ambiguity detection

\textbf{Ambiguity detection}

\begin{verbatim}
This bot determines whether a given question is ambiguous or not.
Question: Tell me about Obama family tree.	
This question is ambiguous: False.

Question: TV on computer.
This question is ambiguous: True.

Question: What is Fickle Creek Farm
This question is ambiguous: False.

Question: Find condos in Florida.
This question is ambiguous: True.

[Three additional examples]

Question: [question to be classified]
This question is ambiguous:
\end{verbatim}

As explained in the main part of the paper, we then take the log probability of the next token being \texttt{True} as a continuous predictor of whether the given question is ambiguous or not.

\subsection{CLAQUA I}

This data set corresponds to the single-turn part of the CLAQUA \citep{xu2019asking} data set. A given sample in this data set consists of the following components: two fields \textit{entity1} and \textit{entity2} that provide descriptions on two entities which both have the same name but are two distinct entities. Each sample also contains a question that is either ambiguous or unambiguous, that is a question that could possibly refer to either of the two entities or not. For the ambiguous question, the data set additionally contains a label that indicates which of the two entities the question is supposed to refer to, as well as an answer to the ambiguous initial question. This data set does not contain answers for the unambiguous questions in the data set.

As done in \citet{xu2019asking}, the description of the two entities is provided to the question answering model together with the given question.

\textbf{Ambiguity detection}

We use the following few-shot prompt to detect whether a given question is ambiguous or not
\begin{verbatim}
    
This bot determines whether a given question is ambiguous or not.

entity1: Aggrenox  Aggrenox (Aspirin, Dipyridamole) (...)
entity2: Aggrenox Aggrenox contains a combination of (...)
"What is Aggrenox's ingredient?" could refer to both entities "Aggrenox": True.

entity1: Jack Shelton Jack Shelton (born November 30, 1931) is an American bebop and(...)
entity2: Jack Shelton John Jack Shelton was an English footballer who played as a right-half and inside-forward. (...)
"What are the written works of Jack Shelton?" could refer to both \\
entities "Jack Shelton": False.

[Three additional examples]

[description of entity1]
[description of entity2]
[question] could refer to both entities [entity name]:

\end{verbatim}

As explained in the main part of the paper, we then take the log probability of the next token being \texttt{True} as a continuous predictor of whether the given question is ambiguous or not.

\textbf{Clarifying question generation}

\begin{verbatim}
Question: Casting director for Fakers
When you say Fakers, are you referring to the TV movie or the movie?

###

Question: What is name of place where Ernest Pollard was born?
When you say Ernest Pollard, are you referring to the Nebraska Republican politician or the professor of physics and biophysics?

###

Question: Cunningham Elementary\'s rank
Which Cunningham Elementary are you referring to?

###

Question: Who is the host of Room 101?
Which Room 101?


###
Question: {initial_question}
\end{verbatim}

\subsection{CLAQUA II}
This data set corresponds to the multi-turn part of the CLAQUA (Xu et al., 2019) data set. Every sample of the data set contains the following components: A three-turn dialogue where the last turn is a question. For some samples, this question can refer to

\textbf{Ambiguity detection}

\begin{verbatim}
This bot determines whether a given question is ambiguous or not.

Context: Bazil Broketail. Bazil Broketail (1992) is a fantasy novel written by (...)
Context: A Sword  for a Dragon. A Sword for a Dragon is a fantasy novel written (...)
User: Is there a sequel to bazil broketail 
Bot: A Sword for a Dragon 
User: How about the style of this creative work?
"How about the style of this creative work?" could refer both \\ 
to Bazil Broketail and A Sword for a Dragon: True

###

Context: 807 Ceraskia.  807 Ceraskia is a minor planet orbiting the Sun.
Context: Eos family.  The Eos family (adj. Eoan; FIN: 606) is a very large asteroid (...)
User: What is family for 807 ceraskia
Bot: Eos family
User: What is the star system of the object
"What is the star system of the object" could refer both \\
to 807 Ceraskia and Eos family: False.

[Three additional examples]

###

[context]
[dialogue]
{question} could refer both to [entity1] and [entity2]:
\end{verbatim}

As explained in the main part of the paper, we then take the log probability of the next token being \texttt{True} as a continuous predictor of whether the given question is ambiguous or not.

\textbf{Clarifying question generation}

\begin{verbatim}
This is a conversation between a user and a question-answering bot.
Is there a sequel to bazil broketail <EOS>\\ 
A Sword for a Dragon <EOS>\\
How about the style of this creative work?
Bot: To answer this question, I need to ask the following clarfiying question:
Are you referring to bazil broketail or a sword for a dragon?

###

What is higher classification for schadonia <EOS>\\
Lecanorineae <EOS>\\
Which cataloge it should be classified into?
To answer this question, I need to ask the following clarfiying question:
Are you referring to schadonia or Lecanorineae?

[Two additional examples]

###

{question}
Bot: To answer this question, I need to ask the following clarfiying question:
\end{verbatim}

\newpage

\section{CLAM Algorithm}

We tune the decision threshold $\tau$ used to detect ambiguity in CLAM (see Algorithm \ref{alg:selective_clarification}) to maximize the QA accuracy on a holdout data set of ambiguous and unambiguous questions and use $\tau = -0.3$ in the remaining experiments. 

\begin{algorithm*}
\caption{Selective clarification of imprecise questions}
\begin{algorithmic}
\REQUIRE{A language model $\mathcal{M}$, a question $\mathcal{Q}$, a user $\mathcal{U}$, ambiguity classifier $f$.}
\STATE $\mathcal{A} \gets \mathcal{M}(\mathcal{Q})$ \COMMENT{Ask language model to answer question $\mathcal{Q}$}
\STATE $\texttt{ambiguous} \gets f(\mathcal{Q}, \mathcal{M})$ \COMMENT{Classify ambiguity, e.g., with few-shot prompted LLM}
\IF{$\texttt{ambiguous}$} 
        \STATE $\mathcal{Q}' \gets concat(\mathcal{Q, \textrm{``To answer this question,
        I have to ask the following clarifying question''}})$
        \STATE $\hat{\mathcal{Q}} \gets \mathcal{M}(Q')$ \COMMENT{Ask a clarifying question $\hat{\mathcal{Q}}$}
        \STATE $\hat{\mathcal{A}} \gets \mathcal{U}(concat(Q, \hat{\mathcal{Q}}))$ \COMMENT{Get clarification from user or oracle} 
        \STATE $\mathcal{A} \gets \mathcal{M}(concat(Q, \hat{\mathcal{Q}}, \hat{\mathcal{A}}))$ \COMMENT{Return answer given entire dialogue}
\ENDIF \\
\STATE $\text{\textbf{return }} \mathcal{A}$ 
\end{algorithmic}
\label{alg:selective_clarification}
\end{algorithm*}


\end{document}